\documentclass[12pt]{article}
\usepackage{amsthm}
\usepackage{amsmath}
\usepackage{amsfonts}
\usepackage{amssymb}
\usepackage{url}
\usepackage{epsfig}

\newtheorem{theorem}{Theorem}[section]

\newtheorem{lemma}[theorem]{Lemma}

\theoremstyle{definition}
\newtheorem{de}[theorem]{Definition}

\newtheorem{remark}[theorem]{Remark}

\newcommand{\R}{\mathbb{R}}

\begin{document}
\bibliographystyle{plain}

\title{Discrete Schemes for Gaussian
Curvature  and  Their Convergence  }
\author{
 {Zhiqiang Xu\thanks{Email:xuzq@lsec.cc.ac.cn} \hspace*{8mm}
 Guoliang Xu \thanks{Email:xuguo@lsec.cc.ac.cn}}
\\
{Institute of Computational Math., }
 \\
 Academy of Mathematics and System Sciences,
 \\
 Chinese
 Academy of Sciences, Beijing, 100080, China
}

\maketitle

\begin{abstract}
In this paper, a new discrete scheme for Gaussian curvature is
presented. We show that this new scheme converges at the regular
vertex with valence not less than 5. By constructing a
counterexample, we also show that it is impossible for building a
discrete scheme for Gaussian curvature which converges over the
regular vertex with valence 4. Moreover, the convergence property
of a modified discrete scheme for the Gaussian curvature on
certain meshes is presented.  Finally, asymptotic errors of
several discrete schemes for Gaussian curvature are compared.
\end{abstract}

\section{Introduction}

Some applications from computer  vision, computer graphics,
geometric modeling and computer aided design require estimating
intrinsic geometric invariants.  It is well known that Gaussian
curvature is one of the most essential geometric invariants for
surfaces. However, in the classical differential geometry, this
invariant is well defined only for $C^2$ smooth surfaces. In modern
computer-related geometry fields, one often uses $C^0$ continuous
discrete triangular meshes to represent smooth surfaces
approximately. Hence, the problem of estimating accurately Gaussian
curvature for triangular meshes is raised naturally.

In the past years, a wealth of different estimations have been
proposed in the vast literature of applied geometry.   These methods
for estimating Gaussian curvature can be  divided into two classes.
The first class is based on the local fitting or interpolation
technique \cite{cazals,douros,martin,dsmeek,xulb2},
 while the second class is based on discretization formulations which
represent the information about the Gaussian curvature
\cite{angular,calladine,dsmeek,meyer,dissurv,xugau}. In this
paper, our focus is on the methods in the second class. The main
aim of the paper is to present a new discrete scheme for the
discrete Gaussian curvature which converges at the regular vertex
with valence not less than 5.

Let $M$ be a triangulation of smooth surface $S$ in ${\R}^3.$ For
a vertex ${\bf p}$ of $M, $ suppose $\{{\bf p}_i\}_{i=1}^n$ is the
set of the one-ring neighbor vertices of ${\bf p}$. The set
$\{{\bf p}_i{\bf p}{\bf p}_{i+1}\}$ $(i=1,\cdots,n)$ of $n$
Euclidean triangles forms a piecewise linear approximation of $S$
around ${\bf p}$. Throughout the paper, we use the following
conventions  ${\bf p}_{n+1}={\bf p}_1$ and ${\bf p}_0={\bf p}_n$.
Let $\gamma_i$ denote the angle $\angle {\bf p}_i{\bf pp}_{i+1}$
and the angular defect at ${\bf p}$ be $2\pi-\sum_i\gamma_i$.

 A popular discrete scheme for computing Gaussian curvature is in the
 form of $\frac{2\pi-\sum_i\gamma_i}{E}$, where $E$ is a geometry
 quantity. In general, one selects $E$ as $A({\bf p})/3$ and obtain the following
 approximation
 \begin{equation}\label{eq:k1}
 G^{(1)}\,\, :=\,\, \frac{3(2\pi-\sum_i\gamma_i)}{A({\bf p})},
 \end{equation}
  where $A( {\bf p})$ is
 the sum of the areas of triangles ${{\bf p}_i{\bf p}{\bf p}_{i+1}}.$ In \cite{angular},
another scheme
\begin{equation}\label{eq:k2}
 G^{(2)}\,\, :=\,\, \frac{2\pi-\sum_i\gamma_i}{S_p}
\end{equation}
  is
given, where
\[S_p\,\, :=\,\, \sum_i\frac{1}{4\sin\gamma_i}\left[\eta_i \eta
_{i+1}-\frac{\cos\gamma_i}{2}(\eta_i^2+\eta_{i+1}^2)\right] \] is
called the module of the mesh at ${\bf p}$. In \cite{dissurv}, the
discrete approximation $G^{(1)}$ is modified as
\begin{equation}\label{eq:k3}
G^{(3)} := \frac{2\pi-\sum_i \gamma_i}{\frac{1}{2}\sum_i {\rm
area}({{\bf p}_i{\bf p}{\bf
p}_{i+1}})-\frac{1}{8}\sum_i\cot(\gamma_i)d_i^2},
\end{equation}
 where $d_i$ is
the length of  edges ${\bf p}_i{\bf p}_{i+1}.$  There are several
different points of view for explaining the reason why the angular
defect is closely related to the Gaussian curvature with  including
the viewpoints of Gaussian-Bonnet theorem, Gaussian map and
Legendre's formula (see the next section for details).

Asymptotic analyses for the discrete schemes have been given in
\cite{angular,dsmeek,xugau}. In \cite{dsmeek}, the authors show
that for the non-uniform data, the discrete scheme $G^{(1)}$ does
not convergent to true Gaussian curvature always. In
\cite{angular}, Borrelli et al. prove that the angular defect is
asymptotically equivalent to a homogeneous polynomial of degree
two in the principal curvatures and show that if ${\bf p}$ is a
regular vertex with valence six, then the scheme $G^{(2)}$
converges to the exact Gaussian curvature in a linear rate.
Moreover, Borrelli et al. show that $4$ is the only value of the
valence such that the angular defect depends upon the principal
directions.   In \cite{xugau}, Xu proves that the discrete scheme
$G^{(1)}$ has quadratic convergence rate if the mesh satisfies the
so-called parallelogram criterion, which requires valence 6.
Therefore, one hopes to construct a discrete scheme which
converges over any discrete mesh. But in \cite{zxu}, Xu et al.
show that it is impossible to construct a discrete scheme which is
convergent for any discrete mesh. Hence, we have to be content
with the discrete schemes which converge under some conditions.
According to the past experience \cite{angular,tlanger,zxu}, we
regard a discrete scheme desirable if it has the following
properties
\begin{enumerate}
\item It converges at a regular vertex, at least for sufficiently large
valence (the definition of the regular vertex will be given in
Section 2);

\item It converges at the umbilical point, i.e., the points satisfying $k_m=k_M$ where $k_m$
and $k_M$ are two principal  curvatures.
\end{enumerate}

As stated before, the previous discrete schemes, including
$G^{(1)}, G^{(2)}$ and $G^{(3)}$, only converge at the regular
vertex with valence 6. In \cite{angular}, a method for computing
the Gaussian curvature at the regular vertex with valence unequal
to 4 is described. But the method requires two meshes with
valences $n_1$ and $n_2$ ($n_1\neq 4,n_2\neq 4,n_1\neq n_2$). In
this paper, we will construct a discrete scheme which converges at
the regular vertex with valence not less than 5, and also at the
umbilical points with any valence. Moreover, the discrete scheme
requires only a single mesh. Hence, the new discrete scheme is
more desirable. Furthermore, we show that it is impossible to
construct a discrete scheme which is convergent at the regular
vertex with valence 4. Therefore, for the regular vertex, the
convergence problem remains open for the case of vertex with
valence 3. We also want to point out that we discuss the pointwise
convergence in this paper, which is different with the convergence
in norm as discussed in \cite{new:morvan,new:hilde}.

The rest of the paper is organized as follows. Section 2 describes
some notations and definitions and  Section 3 shows three
different viewpoints for expressing the relation between the
angular defect and Gaussian curvature. In Section 4,
 we study the convergence property of the modified discrete
Gaussian curvature scheme. We  present in Section 5 a new discrete
scheme and prove that the scheme has good convergence property. In
Section 6, for the regular vertex with valence 4, we show that it is
impossible to build a discrete scheme which is convergent to the
real Gaussian curvature. Some numerical results are given in Section
7.

\section{Preliminaries}

\begin{center}
\epsfxsize=7cm\epsfbox{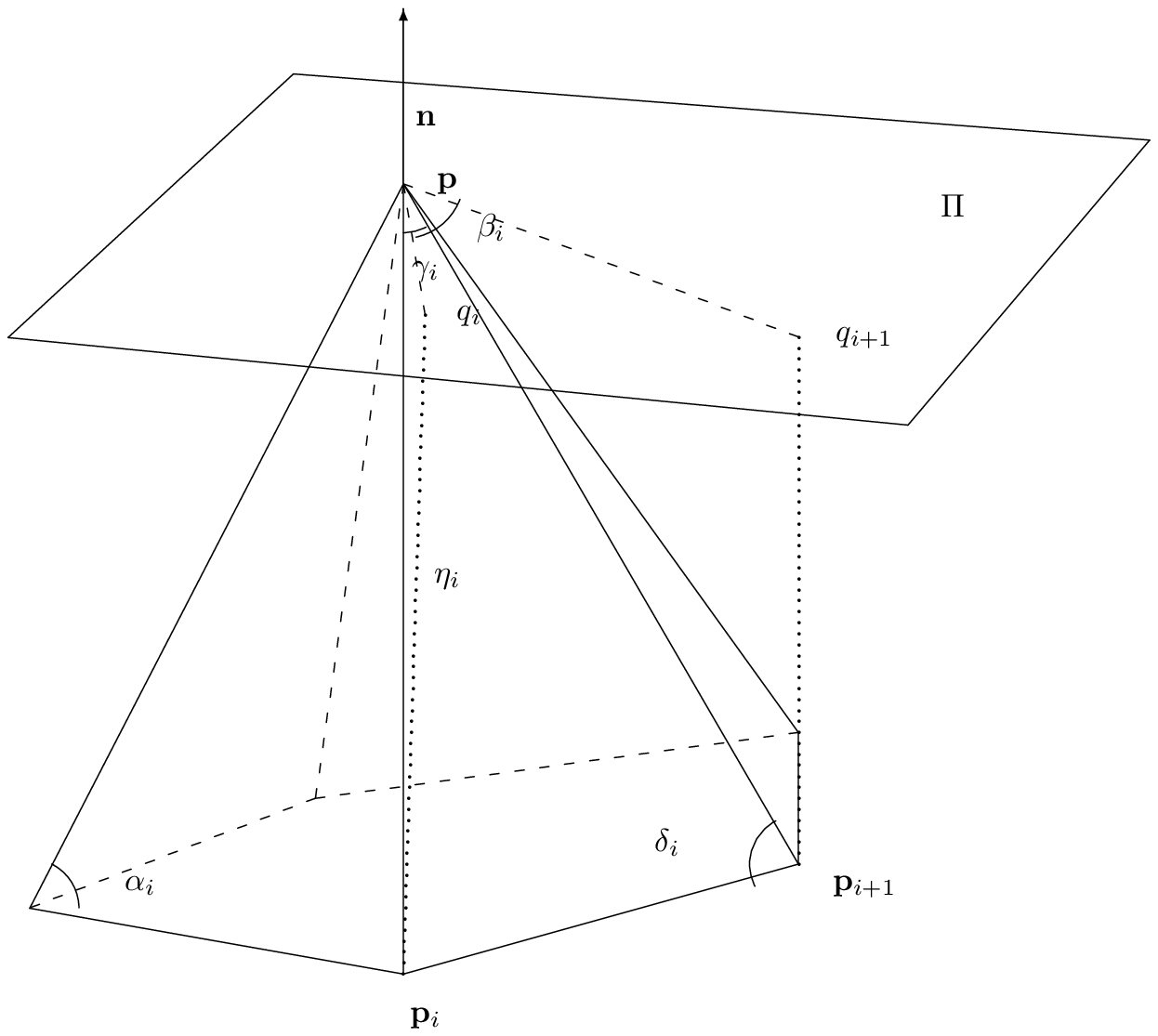}
\end{center}
\begin{center}
{\normalsize {\bf Fig.1.} Notations.}
\end{center}

In this section, we introduce some notations and definitions used
throughout the paper (see Fig. 1). Let $S$ be a given smooth
surface and ${\bf p}$ be a point over $S.$ Suppose the set $\{{\bf
p}_i{\bf pp}_{i+1}\}$, $i=1,\cdots,n$, of $n$ Euclidean triangles
form a piecewise linear approximation of $S$ around ${\bf p}.$ The
vector from ${\bf p}$ to ${\bf p}_i$ is denoted as
$\overrightarrow{{\bf pp}_i}.$ The normal vector and tangent plane
of $S$ at the point ${\bf p}$ is denoted by ${\bf n}$ and $\Pi,$
respectively. We denote the projection of ${\bf p}_i$ onto $\Pi$
as ${\bf q}_i$, and define the plane containing ${\bf n},{\bf p}$
and ${\bf p}_i$ as $\Pi_i.$ Then we let $\kappa_i$ denote the
curvature of the plane curve $S\cap \Pi_i$ at ${\bf p}$. The
distances from ${\bf p}$ to ${\bf p}_i$ and ${\bf q}_i$ are
denoted as $\eta_i$ and $l_i,$ respectively. The angles $ \angle
{\bf p}_i{\bf pp}_{i+1}$ and $\angle {\bf q}_i{\bf p}{\bf
q}_{i+1}$ are denoted as $\gamma_i$ and $\beta_i,$ respectively.
The two principal curvatures at ${\bf p}$ are denoted as $k_m$ and
$k_M.$ Let $\eta=\max_i \eta_i.$ The following results are
presented in \cite{angular,tlanger,xugau}:
\begin{equation}\label{eq:reg}
\frac{l_i}{\eta_i}=1+O(\eta),\,\,\, \beta_i=\gamma_i+O(\eta^2),
\end{equation}
\begin{equation}\label{eq:weight}
\left\| \sum_i w_i \overrightarrow{{\bf pp}_i} \right\|=\sum_i
\frac{w_i\kappa _i \eta_i^2}2+O(\eta^3),
\end{equation}
where $w_i\in {\R}.$

Now we give the definition of the regular vertex using the notations
introduced  above.
\begin{de}\label{de:regu}
Let ${\bf p}$ be a point of a smooth surface $S$ and let ${\bf
p}_i,i=1,\cdots,n$ be its one ring neighbors. The point ${\bf p}$
is called a regular vertex if it satisfies the following
conditions

(1) the $\beta_i=\frac{2\pi}{n},$

 (2) the $\eta_i$s all take the
same value $\eta.$
\end{de}

\begin{remark}

We can replace  (1) in Definition \ref{de:regu} by requiring the
$\gamma_i$ all take the same value. Since
$\beta_i=\gamma_i+O(\eta^2)$, all the results in the paper hold also
for the alternative definition.
\end{remark}

\section{Angular Defect and Gaussian Curvature}

In this section, we summarize three different viewpoints for
expressing the relation between angular defect and Gaussian
curvature. These viewpoints have been described in different
literature \cite{dsmeek,dissurv,xugau}. We collect them there.
Throughout the section, we use $G^{(1)}({\bf p})$ to denote the
discrete Gaussian curvature at ${\bf p}$, which is obtained using
$G^{(1)}.$

\subsection{Gaussian-Bonnet theorem viewpoint}

Let $D$ be a region of surface $S,$ whose boundary consists of
piecewise smooth curves $\Gamma_j$s. Then the local Gaussian-Bonnet
theorem is as follows
$$
\int\!\!\!\int_{D}G(p)dA+\sum_j\int_{\Gamma_j}k_g(\Gamma_j)ds+\sum_j\alpha_j=2\pi,
$$
where $G(p)$ is the Gaussian curvature at $p,$ $k_g(\Gamma_j)$ is
the geodesic curvature of the boundary curve $\Gamma_j$ and
$\alpha_j$ is the exterior angle at the $j$th corner point ${\bf
p}_j$ of the boundary. If all the $\Gamma_j$s are the geodesic
curves, the above formula reduces to
\begin{eqnarray}\label{eq:gaussbonn}
\int\!\!\!\int_{D}G(p)dA=2\pi-\sum_j\alpha_j.
\end{eqnarray}
Let $M$ be a triangulation of surface $S$. For vertex ${\bf p}$ of
valence $n$, each triangle ${\bf p}_i{\bf pp}_{i+1}$ can be
partitioned into three equal parts, one corresponding to each of its
vertices. We let $D$ be the  union of the part corresponding to
${\bf p}$ of triangles ${\bf p}_i{\bf pp}_{i+1}.$ Note that
$\sum_i\gamma_i=\sum_j\alpha_j.$ Assuming $G({ \bf p})$ is a
constant on $D$, and using (\ref{eq:gaussbonn}), we have $ G({ \bf
p})$ can be approximated by $G^{(1)}({\bf p})$.

\subsection{Spherical image viewpoint}

We now introduce another definition of Gaussian curvature. Let $D$
be a small patch of area $A$ including point ${\bf p}$ on the
surface $S$. There will be a corresponding patch of area  $I$ on the
Gaussian map. Gaussian curvature at ${\bf p}$ is the limit of ratio
$\lim_{A\rightarrow 0}\frac{I}{A}.$

Let us consider a discrete version of the definition. The Gaussian
map image, i.e. the spherical image, of the triangle ${\bf p}_i{\bf
pp}_{i+1}$ is the point $\frac{({\bf p}-{\bf p}_i)\times ({\bf
p}-{\bf p}_{i+1})}{\| ({\bf p}-{\bf p}_i)\times ({\bf p}-{\bf
p}_{i+1})\|}.$ Join these points by great circle forming a spherical
polygon on the unit sphere.  The area of this spherical polygon is
$2\pi-\sum_i\gamma_i.$ Same as the above, each triangle is
partitioned into three parts, one corresponding to each vertex. Then
the Gaussian curvature can be approximated by $G^{(1)}({\bf p})$.

\subsection{Geodesic triangles viewpoint}

Let $T= ABC$ be a geodesic triangle on the surface $S$ with angles
$\alpha,\beta,\gamma$ and geodesic edge lengths $a,b,c.$ Let
$A'B'C'$ be a corresponding Euclidean triangle with edge lengths
$a,b,c$ and angles $\alpha',\beta',\gamma'$.  Legendre presents the
following formulation
$$
\alpha-\alpha'={\rm area}(T)\frac{G(A)}{3}+o(a^2+b^2+c^2),
$$
where ${\rm area}(T)$ is the area of the geodesic triangle $ABC,$
$G(A)$ is the Gaussian curvature at $A$.

Using Legendre's formulation for each triangles with ${\bf p}$ as a
vertex, we arrive at the estimating formula  $G^{(1)}({\bf p})$
again.

\section{Convergence of Angular Defect Schemes}

In \cite{xugau}, Xu gives an analysis about the scheme $G^{(1)}$
and proves that the scheme converges at the vertex satisfying
so-called parallelogram criterion. A numerical test shows that the
scheme does not converge at the regular vertex with valence
unequal to 6 and at umbilical points. In \cite{angular}, Borrelli
et. al. give an elegant analysis about the angular defect. They
show that if the vertex ${\bf p}$ is regular, then the angular
deficit is asymptotically equivalent to a homogeneous polynomial
of degree two in the principal curvatures with closed form
coefficients. Moreover, they present another angular scheme
$G^{(2)}:= \frac{2\pi-\sum_i\gamma_i}{S_p}$. In fact, using the
law of cosine, we have
\begin{eqnarray*}
&& \hspace*{-7mm}\frac{1}{2}\sum_i {\rm area}({\bf p}_i{\bf
pp}_{i+1})-\frac{1}{8}\sum_i\cot(
\gamma_i)d_i^2\\
&=&\sum_i\left[\frac{1}{4}\eta_i\eta_{i+1}\sin\gamma_i-\frac{1}{8}\frac{\cos\gamma_i}{\sin\gamma_i}
(\eta_i^2+\eta_{i+1}^2-2\eta_i\eta_{i+1}\cos\gamma_i)\right]\\
&=&\sum_i
\frac{1}{4\sin\gamma_i}\left[\eta_i\eta_{i+1}-\frac{\cos\gamma_i}{2}(\eta_i^2+\eta_{i+1}^2)\right]=S_p.
\end{eqnarray*}
This shows that  $G^{(2)}$ and $G^{(3)}$ are equivalent, which means
these two schemes obtain the same value for the same triangular
mesh.

In \cite{xugau}, the author proves that the discrete scheme
$G^{(1)}$ has quadratic convergence rate under the parallelogram
criterion. In the following theorem, we shall show that  the
discrete scheme $G^{(3)}$ has also quadratic convergence rate under
the same criterion.

\begin{theorem}\label{th:val}
Let {\bf p} be a vertex of $M$ with valence six, and let ${\bf
p}_j,j=1,\cdots,6$ be its neighbor vertices. Suppose ${\bf p}$ and
${\bf p}_j,j=1,\cdots,6$ are on a sufficiently smooth parametric
surface ${\bf F}(\xi_1,\xi_2)\in {\R}^3,$ and there exist ${\bf
u},{\bf u}_j\in {\R}^2$ such that
$$
{\bf p}={\bf F}({\bf u}),\,\, {\bf p}_j={\bf F}({\bf u}_j) \mbox{
and } {\bf u}_j-{\bf u}=({\bf u}_{j-1}-{\bf u})+({\bf
u}_{j+1}-{\bf u}),\,\, j=1,\cdots,6.
$$
Then
$$
\frac{2\pi-\gamma_i}{\frac{1}{2}A({\bf p},r)-\frac{1}{8}\sum_i\cot(
\gamma_i(r) )d_i^2(r)}=G({\bf p})+O(r^2),
$$
where, $G({\bf p})$ is the real Gaussian curvature of ${\bf F}({\bf
u})$ at ${\bf p}$,
\[ A({\bf p},r):=\sum_i {\rm area}[{\bf p}_i(r){\bf pp}_{i+1}(r)], \ \
{\bf p}_i(r):={\bf F}({\bf u}_i(r)),\] and ${\bf u}_i(r)={\bf
u}+r({\bf u}_i-{\bf u}),i=1,\cdots,6$.
\end{theorem}
\begin{proof}
  Let
\begin{equation}
A({\bf p},r)=a_0r^2+a_1r^3+O(r^4)\\
\end{equation}
 and
$$
\frac{A({\bf p},r)}{2}-\frac{1}{8}\sum_i \cot(
\gamma_i(r))d_i^2(r)=b_0r^2+b_1r^3+O(r^4)
$$
be the Taylor expansions with respect to $r$. According to Theorem
4.1 in \cite{xugau},
$$
\frac{3(2\pi-\gamma_i)}{A({\bf p},r)}=G({\bf p})+O(r^2).
$$
Hence, to prove the theorem, we need to show
$b_0=a_0/3,b_1=a_1/3$. According to \cite{xugau}, we have $a_1=0$,
which implies that we only need to prove $b_0=a_0/3,b_1=0$.

Note that the ${\bf u},{\bf u}_j,j=1,\ldots 6,$ satisfy the
parallelogram criterion. Without loss of generality, we may assume
${\bf u}=[0,0]^T, {\bf u}_1 = [1, 0]^T.$ Then there exists a
constant $a > 0$ and an angle $\theta$ such that
$$
{\bf u}_2 = [a \cos\theta,a \sin\theta]^T.
$$
Hence, ${\bf u}_3=[a\cos \theta-1,a\sin\theta ]^T,{\bf
u}_{j+3}=-{\bf u}_j,j=1,2,3.$ Let
$$
{\bf u}_j=s_j {\bf d}_j=s_j[g_j,l_j]^T,\,\, j=1,\cdots,6,
$$
where $s_j=\|{\bf u}_j\|$ and $\|{\bf d}_j\|=1$. Then, we have
\begin{eqnarray*}
& &s_1=1,\, s_2=a,\, s_3=\sqrt{a^2-2ac+1},\, s_4=s_1,\, s_5=s_2,\,
s_6=s_3,\\
& & g_1=1,\, g_2=c,\, g_3=(ac-1)/s_3,\, g_4=-g_1,\,
g_5=-g_2,\, g_6=-g_3, \\
& & l_1=0,\, l_2=t,\, l_3=at/s_3,\, l_4=-l_1,\, l_5=-l_2,\,
l_6=-l_3,
\end{eqnarray*}
where $(c,t):=(\cos\theta,\sin\theta)$.  Note that {
\begin{eqnarray}\label{eq:proof1}
A({\bf p},r)=\frac{1}{2}\sum_{j=1}^6 \sqrt{\|{\bf p}_j(r)-{\bf
p}\|^2\|{\bf p}_{j+1}(r)-{\bf p}\|^2-\langle{\bf p}_j(r)-{\bf
p},{\bf p}_{j+1}(r)-{\bf p}\rangle^2},
\end{eqnarray}
\begin{equation}\label{eq:proof2}
\cot (\gamma_j(r))=\frac{\langle{\bf p}_j(r)-{\bf p},{\bf
p}_{j+1}(r)-{\bf p}\rangle}{\sqrt{\|{\bf p}_j(r)-{\bf p}\|^2\|{\bf
p}_{j+1}(r)-{\bf p}\|^2-\langle{\bf p}_j(r)-{\bf p},{\bf
p}_{j+1}(r)-{\bf p}\rangle^2}},
\end{equation}
\begin{equation}\label{eq:proof3}
d_j^2(r)=\|{\bf p}_j(r)-{\bf p}\|^2+\|{\bf p}_{j+1}(r)-{\bf
p}\|^2-2\langle{\bf p}_j(r)-{\bf p},{\bf p}_{j+1}(r)-{\bf p}\rangle.
\end{equation}
}

Let ${\bf F}^k_{{\bf d}_j}$ denote the $k$th order directional
derivative of {\bf F} in the direction ${\bf d}_j.$ Then using
Taylor expansion with respect to $r$, we have { 
\begin{eqnarray}\label{eq:p1}
 \|{\bf p}_j(r)-{\bf p}_j\|^2=s_j^2r^2\langle{\bf F}_{d_j},{\bf
F}_{d_j}\rangle+ s_j^3r^3\langle{\bf F}_{d_j},{\bf
F}_{d_j}^2\rangle+\frac{1}{4}s_j^4r^4 \langle{\bf F}_{d_j}^2,{\bf F}_{d_j}^2\rangle\nonumber\\
+\frac{1}{3}s_j^4r^4 \langle{\bf F}_{d_j},{\bf F}_{d_j}^3\rangle
+\frac{1}{6}s_j^5r^5 \langle{\bf F}_{d_j}^2,{\bf
F}_{d_j}^3\rangle+\frac{1}{12}s_j^5r^5 \langle{\bf F}_{d_j},{\bf
F}_{d_j}^4\rangle+O(r^6),
\end{eqnarray}}
and
\begin{eqnarray}\label{eq:p2}
& &\langle{\bf p}_j(r)-{\bf p},{\bf p}_{j+1}(r)-{\bf
p}\rangle\nonumber\\
&=& s_js_{j+1}r^2\langle{\bf F}_{d_j},{\bf
F}_{d_{j+1}}\rangle+\frac{1}{2}s_js_{j+1}^2r^3\langle{\bf
F}_{d_j},{\bf
F}_{d_{j+1}}^2\rangle+\frac{1}{2}s_j^2s_{j+1}r^3\langle{\bf
F}_{d_{j+1}},{\bf
F}_{d_j}^2\rangle\nonumber\\
&+&\frac{1}{4}s_j^2s_{j+}^2r^4\langle{\bf F}_{d_{j+1}}^2,{\bf
F}_{d_j}^2\rangle+\frac{1}{6}s_js_{j+}^3r^4\langle{\bf F}_{d_j},{\bf
F}_{d_{j+1}}^3\rangle+\frac{1}{6}s_j^3s_{j+1}r^4\langle{\bf
F}_{d_{j+1}},{\bf
F}_{d_{j+1}}\rangle\nonumber\\
&+&\frac{1}{12}s_j^2s_{j+1}^3r^5\langle{\bf F}_{d_j}^2,{\bf
F}_{d_{j+1}}^3\rangle+\frac{1}{12}s_{j+1}^2s_{j}^3r^5\langle{\bf
F}_{d_j}^2,{\bf
F}_{d_{j+1}}^3\rangle\nonumber\\
&+&\frac{1}{24}s_{j+1}^4s_{j}r^5\langle{\bf F}_{d_j},{\bf
F}_{d_{j+1}}^4\rangle +\frac{1}{24}s_{j+1}s_{j}^4r^5\langle{\bf
F}_{d_j}^4,{\bf F}_{d_{j+1}}\rangle+O(r^6).
\end{eqnarray}
 To compute all the inner products in the two equations above, we
let
$$
{\bf t}_i=\frac{\partial {\bf F}(\xi_1,\xi_2)}{\partial\xi_i},\,
{\bf t}_{ij}=\frac{\partial^2 {\bf
F}(\xi_1,\xi_2)}{\partial\xi_i\xi_j},\, {\bf
t}_{ijk}=\frac{\partial^3 {\bf F}}{\partial \xi_i\partial
\xi_j\partial \xi_k},\, {\bf t}_{ijkl}=\frac{\partial^4 {\bf
F}}{\partial \xi_i\partial \xi_j\partial \xi_k\xi_l}
$$ for
$i,j,k,l=1,2$ and
$$
g_{ij}={\bf t}_i^T{\bf t}_j,\, g_{ijk}={\bf t}_i^T{\bf t}_{jk},\,
e_{ijkl}={\bf t}_i^T{\bf t}_{jkl},\, e_{ijklm}={\bf t}_i^T{\bf
t}_{jklm},\, f_{ijklm}={\bf t}_{ij}^T{\bf t}_{klm}.
$$
Since ${\bf F}_{{\bf d}_j}^k$ can be written as the linear
combinatorics of ${\bf t}_i,{\bf t}_{ij},{\bf t}_{ijk}$ and ${\bf
t}_{ijkl},$ all the inner products in (\ref{eq:p1}) and
(\ref{eq:p2}) can be expressed as linear combinations of $g_{ij},
g_{ijk},g_{ijkl},e_{ijkl}, e_{ijklm}$ and $f_{ijklm}.$

Substituting (\ref{eq:p1}) and (\ref{eq:p2}) into (\ref{eq:proof1}),
(\ref{eq:proof2}) and (\ref{eq:proof3}), and then substituting
(\ref{eq:proof1}), (\ref{eq:proof2}) and (\ref{eq:proof3})  into the
expression $\frac12 A({\bf p},r)-\frac{1}{8}\sum_i \cot(
\gamma_i(r))d_i^2(r)$, and using Maple to conduct all the symbolic
calculation, we have
$$
b_0=a_0/3=\sqrt{a^2t^2(g_{11}g_{22}-g_{12}^2)},\,\, b_1=0.
$$
The theorem is proved.
\end{proof}

\begin{remark}
The calculation of $b_0,b_1$ involves a huge number of terms. It
is almost impossible to finish the derivation by hand.  Maple
completes all the computation in 26 seconds on a PC equipped with
a 3.0GHZ Intel(R) CPU. The Maple code that conducts all derivation
of the theorem is available in {\tt
http://lsec.cc.ac.cn/$\sim$xuzq/maple.html}. The interested
readers are encouraged to perform the computation.
\end{remark}

\begin{remark}
It should be pointed out that there is another discrete scheme
\[G^{(4)}:=\frac{2\pi-\sum_i\gamma_i}{A_M({\bf p})},\]
where $A_M({\bf p})$ is the area of Voronoi region. Since
$\sum_i{\rm area}({\bf p}_i{\bf pp}_{i+1})$ could be approximated by
$3A_M(p)$ under some conditions, for example the conditions of
Theorem \ref{th:val}, $G^{(4)}$ is easily derived from $G^{(1)}.$
\end{remark}

\section{A New Discrete Scheme of the Gaussian Curvature and Its Convergence}

In this section, we  introduce a new discrete scheme for Gaussian
curvature  which converges over the umbilical points and regular
vertices with valence greater than  4. This is the main result of
the paper. We firstly discuss some properties about the discrete
mean curvature.
  Setting $\alpha_i=\angle {\bf p}_i{\bf p}_{i-1}{\bf p}$ and $\delta_i=\angle {\bf p}_i{\bf p}_{i+1}{\bf p}$,
   we let  \begin{eqnarray}\label{eq:dism}
H^{(1)}\,\,:=\,\, 2\left\| \frac{\sum_i(\cot
\alpha_i+\cot\delta_i)\overrightarrow{\bf pp}_i} {\sum_i(\cot
\alpha_i+\cot\delta_i)\eta_i^2}\right\|,
\end{eqnarray}
 which is  a popular discrete
scheme for the mean curvature at vertex ${\bf p}$ (c.f.
\cite{pinkall}). Moreover, the real mean curvature and the real
Gaussian curvature at ${\bf p}$ are denoted as $H$ and $G$
respectively.  Then, we have

\begin{lemma}\label{th:dism}
At the regular vertex ${\bf p}$, or the umbilical points, the
discrete scheme $H^{(1)}$ converges linearly to the mean curvature
$H$ as $\eta=\eta_i\rightarrow 0.$
\end{lemma}
\begin{proof}
 Firstly, let us consider the convergence property at
the regular vertex. Since ${\bf p}$ is a regular vertex,
$\frac{\cot\alpha_i+\cot\delta_i}{\cot\alpha_j+\cot\delta_j}=1+O(\eta^2),$
for any different $i$ and $j.$ It follows from equation
(\ref{eq:weight}), we have
$$
\left\| \sum_i(\cot\alpha_i+\cot\delta_i)\overrightarrow{\bf
pp}_i\right\|=\sum_i \frac{(\cot\alpha_i+\cot\delta_i) \eta_i^2k_i}2
+O(\eta^3).
$$
Hence,
$$
H^{(1)}=\sum_i\frac{(\cot\alpha_i+\cot\delta_i)\eta_i^2}{\sum_j(\cot\alpha_j+\cot\delta_j)\eta_j^2}
{\kappa_i}+O(\eta) =\frac{1}{n}\sum_i\kappa_i+O(\eta)=H+O(\eta).
$$

Secondly, we study the convergence properties at  the umbilical
points. Over the umbilical points, $k_i=k_j=H$ for any $i$ and $j.$
Hence, \begin{eqnarray*} H^{(1)}&:=& 2\left\| \frac{\sum_i(\cot
\alpha_i+\cot\delta_i)\overrightarrow{ \bf pp}_i} {\sum_i(\cot
\alpha_i+\cot\delta_i)\eta_i^2}\right\|\\
&=&\frac{\sum_i (\cot\alpha_i+\cot\delta_i) \eta_i^2k_i
+O(\eta^3)}{\sum_i(\cot \alpha_i+\cot\delta_i)\eta_i^2} =H+O(\eta).
\end{eqnarray*}
Combining the two results above, the theorem holds.
\end{proof}

Now, we turn to a new discrete scheme for Gaussian curvature.
 Let $\varphi_i:=\sum_{j=1}^i\gamma_j$ and
$$
G^{(5)}:=\frac{2\pi-\sum_i\gamma_i-2(S_p-A)(H^{(1)})^2}{2A-S_p},
$$
where{
\begin{eqnarray*}
A&:=&\sum_i\frac{1}{4\sin\gamma_i}(\frac{\eta_i\eta_{i+1}}{2}(1-\cos2\varphi_i\cos2\varphi_{i+1})\\
& &\,\,\,-
\frac{\cos\gamma_i}{4}(\eta_i^2\sin^2\varphi_i+\eta_{i+1}^2\sin^2\varphi_{i+1})),
\end{eqnarray*}
 \[ S_p:=\sum_i\frac{1}{4\sin(\gamma_{i})}\left[\eta_i\eta_{i+1}
-\frac{\cos(\gamma_{i})}{2}(\eta_i^2+\eta_{i+1}^2)\right].
\]
}
Then, we have
\begin{theorem}\label{th:disg}
For the regular vertices with valence not less than $5,$ or the
umbilical points,  $G^{(5)}$ converges towards the  Gaussian
curvature $G$ as $\eta_i\rightarrow 0.$
\end{theorem}
\begin{proof}
 We firstly consider the regular vertex case. We set
$\theta(n):=\frac{2\pi}{n}.$ Since ${\bf p}$ is a regular vertex,
$\gamma_{i}=\theta(n)+O(\eta^2)$ for any $i$ according to
(\ref{eq:reg}). After a brief calculation, we have
$A=A'+O(\eta^4),S_p=S_p'+O(\eta^4),$ where
\begin{eqnarray*}
 A'&=&\frac{1}{16\sin
(\theta(n))}[2n-n\cos(2\theta(n))-n\cos(\theta(n))]\eta^2,\\
 S_p'&=&\frac{n}{4\sin(\theta(n))}[1 -{\cos(\theta(n))}]\eta^2.
\end{eqnarray*}
 Hence, we have
\begin{eqnarray*}
& &{(2\pi-\sum_i\gamma_i-2(S_p-A)(H^{(1)})^2)}/{(2A-S_p)}\\
&=&{(2\pi-\sum_i\gamma_i-2(S'_p-A')(H^{(1)})^2)}/(2A'-S'_p)+O(\eta^2).
\end{eqnarray*}
 Note that
$\frac{\eta_{max}}{\eta_{min}}=1+O(\eta).$ According to Theorem 3
in \cite{angular}, we have
$$
2\pi-\sum_i\gamma_i=A'G+B'(k_M^2+k_m^2)+o(\eta^2),
$$
where, $ B'=\frac{1}{16\sin(\theta(n))}[n +\frac{n}{2}\cos
(2\theta(n))-\frac{3n}{2}\cos(\theta(n))]\eta^2.$

Note that  $S'_p=A'+2B'$ and
\begin{eqnarray*}
A'G+B'(k_M^2+k_m^2)&=&A'G+B'[(k_M+k_m)^2-2k_Mk_m]\\[1mm]
&=&A'G+4B'H^2-2B'G\\[1mm]
&=&(A'-2B')G+4B'H^2.
\end{eqnarray*}
Hence, $2\pi-\sum_i\gamma_i=(A'-2B')G+4B'H^2+o(\eta^2).$
 Note that $A'=O(\eta^2),B'=O(\eta^2)$ and  $A'-2B'\neq 0$ provided $n\neq 3.$
According to Lemma \ref{th:dism}, $H^{(1)}$ converges to the real
mean curvature.
 Hence, we have,
 when $n\geq 5,$
\begin{eqnarray*}
G&=&\frac{2\pi-\sum_{i}\gamma_i-4B'H^2}{A'-2B'}+o(1)\\
& =&
\frac{2\pi-\sum_i\gamma_i-2(S'_p-A')(H^{(1)})^2}{2A'-S'_P}+o(1)=G^{(5)}+o(1).
\end{eqnarray*}

 Therefore, $G^{(5)}$  converges to the
 Gaussian curvature.

Now, let us consider  the umbilical point case. For  umbilical
points, each directional is the principal direction. According to
Lemma 4 in \cite{angular}, we have
$$
2\pi-\sum_i\gamma_i=(AG+(S_p-A)k_m^2)+o(\eta^2)
$$
 over the umbilical points. Since $k_m^2=H^2=G,$ we have
\begin{eqnarray*}
2\pi-\sum_i\gamma_i&=&(AG+(S_p-A)k_m^2)+o(\eta^2)\\
&=&(AG+2(S_p-A)H^2-(S_p-A)G)+o(\eta^2).
\end{eqnarray*}
Hence,
$$
G=\frac{2\pi-\sum_i\gamma_i-2(S_p-A)(H^{(1)})^2}{2A-S_p}+o(1)=G^{(5)}+o(1).
$$
The theorem holds.
\end{proof}
\begin{remark}
Theorem \ref{th:disg} shows that the new scheme $G^{(5)}$ converges
over the regular vertex with valence greater than 4. As shown
before, the previous schemes only converge over the regular vertex
with valence 6, and hence the new scheme has better convergence
properties over the available scheme.
\end{remark}

\begin{remark}
In \cite{tlanger}, the authors also prove  that the discrete
scheme $H^{(1)}$ converges to the real mean curvature at the
regular vertex. However, the definition of the regular vertex in
\cite{tlanger} is different with our definition.
\end{remark}

\begin{remark}
According to the conclusions above, the Gaussian curvature and mean
curvature can be approximated over the regular vertex with valence
greater than 4. Hence, using the formulation $k_m=H-\sqrt{H^2-G},
k_M=H+\sqrt{H^2-G},$ one can approximate the principal curvatures
over the regular vertex with valence greater than 4.
\end{remark}

\section{A Counterexample for the Regular Vertex with Valence 4}

In \cite{zxu}, we have constructed a triangular mesh and shown that
it is impossible to construct a discrete Gaussian curvature scheme
which converges for that mesh. But the vertex in the mesh is not
regular. In this section, we shall show that it is also impossible
to build a discrete Gaussian curvature scheme which converges over
the regular vertex with valence 4.

Suppose the $xy$ plane  is triangulated around $(0,0)$ by choosing
4 points ${\bf q}_1= (r_1,0),{\bf q}_2=(0,r_1),{\bf q}_3=(-r_1,0)$
and ${\bf q}_4=(0,-r_1)$.  For a bivariate function $f(x,y),$ the
graph of $f(x,y)$, i.e. ${\bf F}(x,y)=[x,y$, $f(x,y)]^T$, can be
regarded as a parametric surface. Let ${\bf p}_0={\bf F}(0,0)$ and
${\bf p}_i={\bf F}({\bf q}_i),i=1,2,3,4.$ The set of triangles
${\bf p}_i{\bf p}_0{\bf p}_{i+1}$ forms a triangular mesh
approximation of ${\bf F}$ at ${\bf p}_0.$ The triangular mesh is
denoted as $M_f.$ When $f(x,y)$ is in the form of $x^2+cxy+y^2$
where $c\in {\R},$ it is easy to prove that ${\bf p}_0:=(0,0,0)^T$
is a regular vertex with valence 4. Moreover, ${\bf
p}_1=(r_1,0,r_1^2)^T, {\bf p}_2=(0,r_1,r_1^2)^T,{\bf
p}_3=(-r_1,0,r_1^2)^T,{\bf p}_4=(0,-r_1,r_1^2)^T.$ Now we show
that it is impossible to construct a discrete scheme for Gaussian
curvature which converges over the vertex ${\bf p}_0$ (See Fig.
2).

We assume that the discrete scheme for Gaussian curvature involving
one-ring neighbor vertices of ${\bf p}_0,$ which is denoted as
$G(M_f,{\bf p}_0;$ ${\bf p}_1,{\bf p}_2,{\bf p}_3,{\bf p}_4),$ is
convergent for the regular vertex with valence 4  over triangular
mesh surface $M_f,$ where $f(x,y)$ is in the form of $x^2+cxy+y^2$.

It is easy to calculate that the  Gaussian curvature of ${\bf
F}(x,y,z)$ at $p_0$ is $4-c^2.$ By the convergence property of
$G(M_f,{\bf p}_0;{\bf p}_1,{\bf p}_2,{\bf p}_3,{\bf p}_4)$  we
have $\lim_{r_1\rightarrow 0}G(M_f,{\bf p}_0;$ ${\bf p}_1,{\bf
p}_2,{\bf p}_3,{\bf p}_4)=4-c^2.$ Note that the triangular mesh
$M_f$ is independent of $c$, i.e. for any function $f(x,y)$ which
is in the form of $x^2+cxy+y^2$, the triangular mesh $M_f$ is the
same. Hence, $\lim_{r_1\rightarrow 0}G(M_f,p_0;p_1,p_2,p_3,p_4)$
is independent of $c.$ A contradiction occurs.

Hence, the assumption of $G(M_f,p_0;p_1,p_2,p_3,p_4)$ being
convergent for the triangular mesh $M_f$ does not hold.

\begin{center}
\epsfxsize=2.5cm\epsfbox{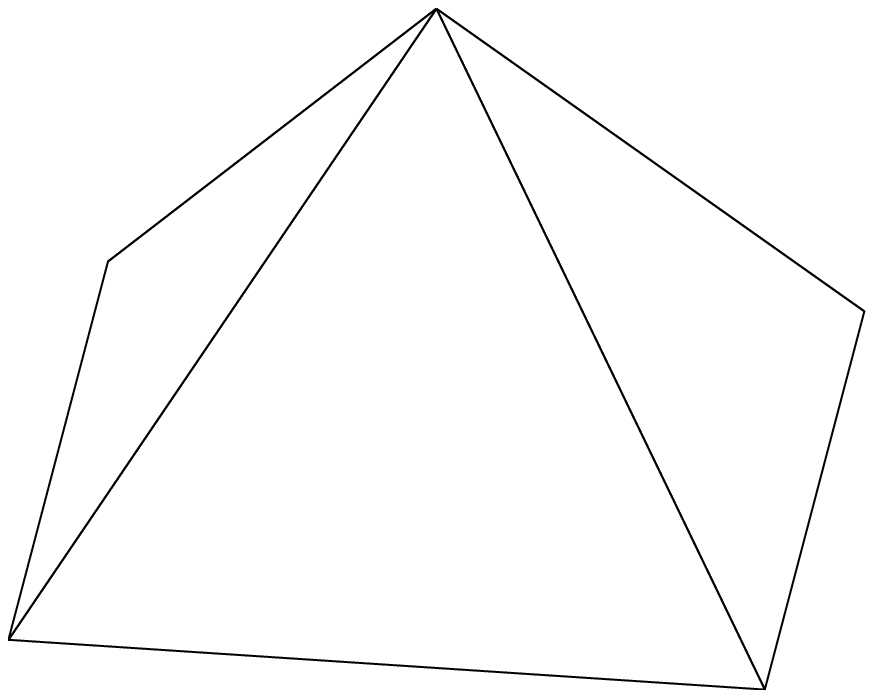}
\epsfxsize=2.5cm\epsfbox{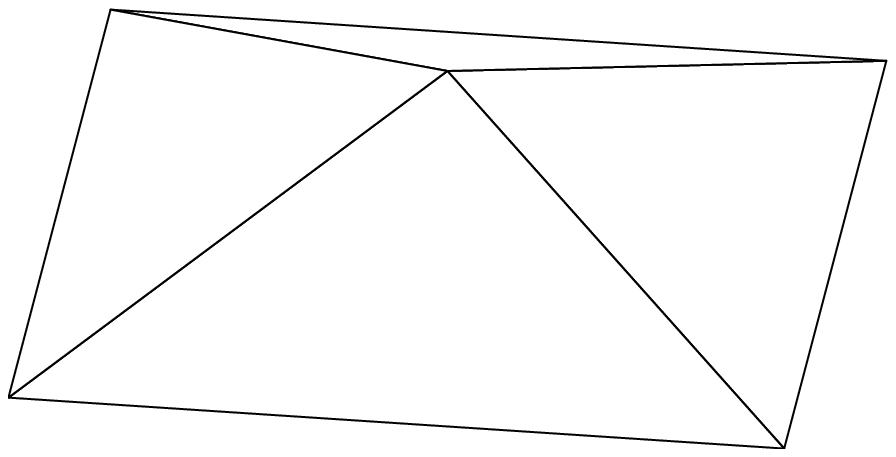}
\epsfxsize=2.5cm\epsfbox{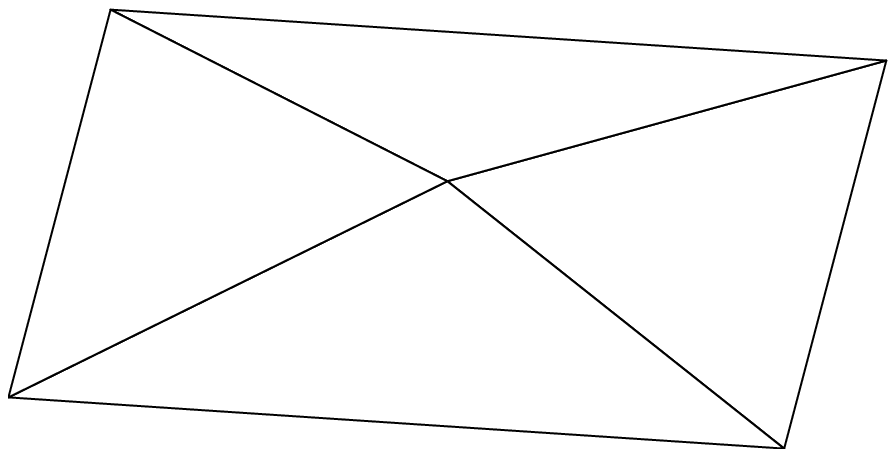}
\epsfxsize=2.5cm\epsfbox{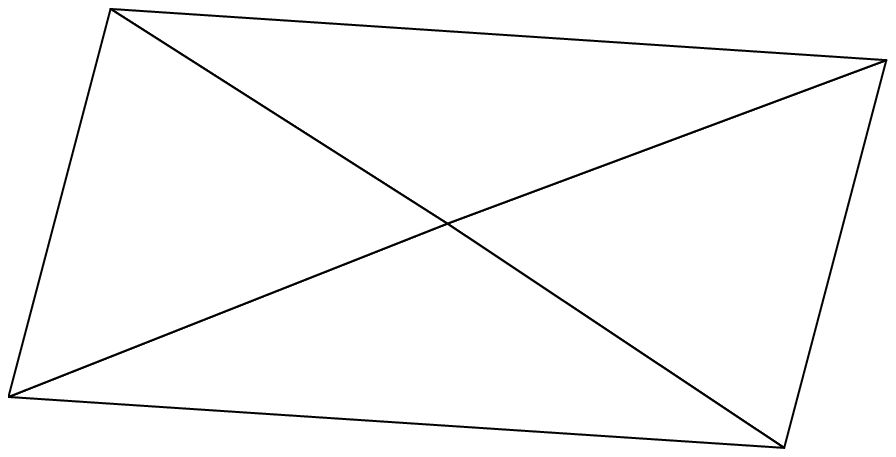}
\end{center}
{\normalsize  {\bf Fig.2.} A sequence of regular vertex with valence
$n=4$ for the function $f(x,y)=x^2+xy+y^2.$ At the regular vertex,
it is impossible to construct a discrete Gaussian curvature scheme
which converges to the correct value.}

\begin{remark}
The counterexample in this section justifies the conclusion in
\cite{angular}, which says that 4 is the only value of valence
such that $2\pi-\sum_i\gamma_i$ depends upon the principal
directions.
\end{remark}

\begin{remark}
 An open problem is to find a discrete scheme for Gaussian curvature
which converges at the regular vertex with valence 3.

\end{remark}

\section{Numerical Experiments}

The aim of this section is to exhibit the numerical behaviors of the
discrete schemes mentioned above. For a real vector ${\bf
a}=(a_{20},a_{11},a_{02}),$ we define a bivariate function $f_{\bf a
}(x,y):=a_{20}x^2+a_{11}xy+a_{02}y^2,$ and regard the graph of the
function $f_{\bf a}(x,y)$ as a parametric surface
$$
{\bf F_a}(x,y)=[x,y,f_{\bf a}(x,y)]^T\in {\R}^3.
$$
The Gaussian curvature  of ${\bf F}_{\bf a}(x,y)$ at the origin is
$4a_{20}a_{02}-a_{11}^2.$ The domain around $(0,0)$ is
triangulated locally by choosing $n$  points:
$$
{\bf q}_k=l_k (\cos \theta_k,\sin \theta_k),\,\,\,
\theta_k=2(k-1)\pi/n,\,\,\, k=1,\cdots,n.
$$

 Let ${\bf p}_k={\bf F}_{\bf a}({\bf q}_k)$ and ${\bf
p}_0=(0,0,0)^T$.  Hence, the set of triangles $\{{\bf p}_k {\bf p}_0
{\bf p}_{k+1}\}$ forms a piecewise linear approximation of ${\bf
F_a}$ around ${\bf p}_0.$ We set $e_k:=f_a(\cos\theta_k, \sin
\theta_k)$ and select
\begin{equation}\label{eq:lk}
l_k=\sqrt{\frac{\sqrt{1+4e_{k}^2(l_{k-1}^2+l_{k-1}^4e_{k-1}^2)}-1}{2e_{k}^2}},\,\,
k\geq 2
\end{equation}
so that ${\bf p}_0$ is a regular vertex.

We let $G^{(i)}({\bf p}_0:{\bf F_a})$ denote the approximated
Gaussian curvatures  of ${\bf F_a}$ at ${\bf p}_0$, which is
obtained by using the discrete scheme $G^{(i)}$. Suppose
${\mathcal A}$ is a set consisting of $M$ randomly chosen vectors
${\bf a}.$ Then, we let
$$
\varepsilon^{(i)}(n)= \sum_{{\bf a}\in {\mathcal A}}|G^{(i)}({\bf
p}_0:{\bf F_a})-(4a_{20}a_{02}-a_{11}^2)|/M.
$$

In fact, $\varepsilon^{(i)}(n)$ measures the error of the discrete
scheme $G^{(i)}$ at the regular vertex with  valence $n.$ The
convergence property and the convergence rate are checked by taking
$l_1=1/8,1/16,1/32,\cdots $. (when $k\geq 2$, $l_k$ can be obtained
by (\ref{eq:lk}).) Since ${\bf p}$ is regular, each edge has the
same length $\eta.$ Table 1 shows the asymptotic maximal error
$\varepsilon^{(i)}(n)$ for $M=10^4.$ Here, the vertex valences $n$
are taken to be $4,5,\cdots,8.$

\begin{table*}[!t]
 \caption{The asymptotic maximal error
$\varepsilon^{(i)}(n).$}
\centerline{
\begin{tabular}{c|c|c|c|c}
\hline
n & $\varepsilon^{(1)}(n)$ & $\varepsilon^{(2)}(n)$ & $\varepsilon^{(4)}(n)$ & $\varepsilon^{(5)}(n)$ \\
\hline
4 & $4.6016e+01$ & $3.3571e+01$& $3.3570e+01$ & $3.3593e+01$ \\
5 & $8.2000e+00$ & $9.3792e+00$ & $9.3792e+00$ & $4.1631e+01\eta$ \\
6 & $1.2226e+01\eta$ & $1.2903e+01\eta$ & $1.2903e+01\eta $ & $1.1488e+01\eta$\\
7 & $3.8464e+00 $ & $4.5783e+00$ & $4.5783e+00$ & $9.0676e-01\eta$\\
8 & $5.8387e+00$ & $7.7628e+00$ & $7.7628e+00$ & $6.5630e+01\eta$\\
\hline
\end{tabular}
}
\end{table*}
From  table 1, we can see that all methods work well on valence 6
but only new method works well for valence $\geq 5$.

 We compute the Gaussian curvature over a
randomly triangulated unit sphere by the discrete schemes to test
their convergent property at the umbilical points. The vertexes of
the random triangulation are uniform distribution on the sphere.
Fig. 3 shows the random triangulation for the unit sphere. Denote
the vertices in the random triangulation as ${\bf
p}_i,i=1,\cdots,N$ where $N$ is the number of the vertices in the
random triangulation. We let $G^{(j)}({\bf p}_i)$ denote the
approximate Gaussian curvature at the vertex ${\bf p}_i$ which is
calculated by $G^{(j)}.$ Similarly to the above, we use $
\varepsilon^{(j)}= \sum_{i=1}^N|G^{(j)}({\bf p}_i)-1)|/N $ to
measure the error of discrete scheme $G^{(j)}$ and use $\eta$ to
denote the average length of the edges. Table 2 lists
$\varepsilon^{(j)}$ for different $N.$ Moreover, we also use
$\varepsilon^{(H)}$ to denote the error of the discrete scheme
$H^{(1)}$ for the mean curvature.

\begin{table*}[h!]
\caption{The asymptotic  error $\varepsilon^{(i)}$ over a sphere
with very irregular connectivity.}

\centerline{
\begin{tabular}{c| c| c| c| c| c|c}
\hline
N & $\eta$  & $\varepsilon^{(1)}$ & $\varepsilon^{(2)}$ & $\varepsilon^{(4)}$ & $\varepsilon^{(5)}$ & $\varepsilon^{(H)}$ \\
\hline
30& 0.710 & $3.798e-01$ & $1.905e-01$& $1.905e-01$ & $2.126e-01$ &2.840e-02\\
100&0.383 & $3.517e-01$ & $5.480e-02$ & $5.480e-02$ & $1.192e-01$ &1.301e-02\\
400&0.196 & $2.673e-01$ & $1.280e-02$ & $1.280e-02 $ & $1.730e-02$&2.600e-03\\
1300&0.109 & $2.812e-01$ & $3.801e-03$ & $3.801e-03$ & $6.500e-03$&7.540e-04\\
5000&0.056 & $2.669e-01$ & $9.648e-04$ & $9.648e-04$ & $2.703e-03$&1.893e-04\\
\hline
\end{tabular}
}
\end{table*}

\begin{center}
\epsfxsize=2.3cm\epsfbox{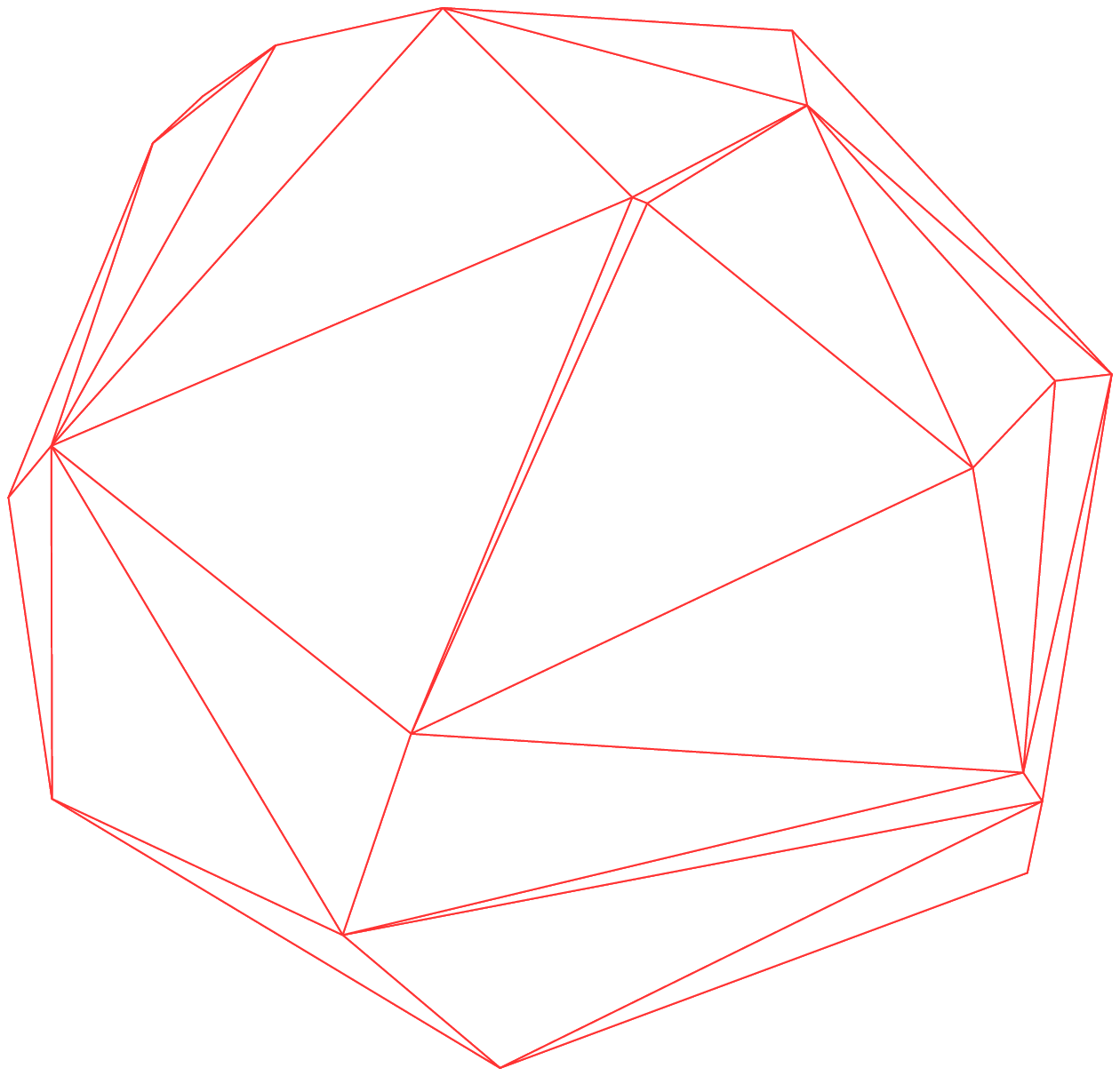}
\epsfxsize=2.3cm\epsfbox{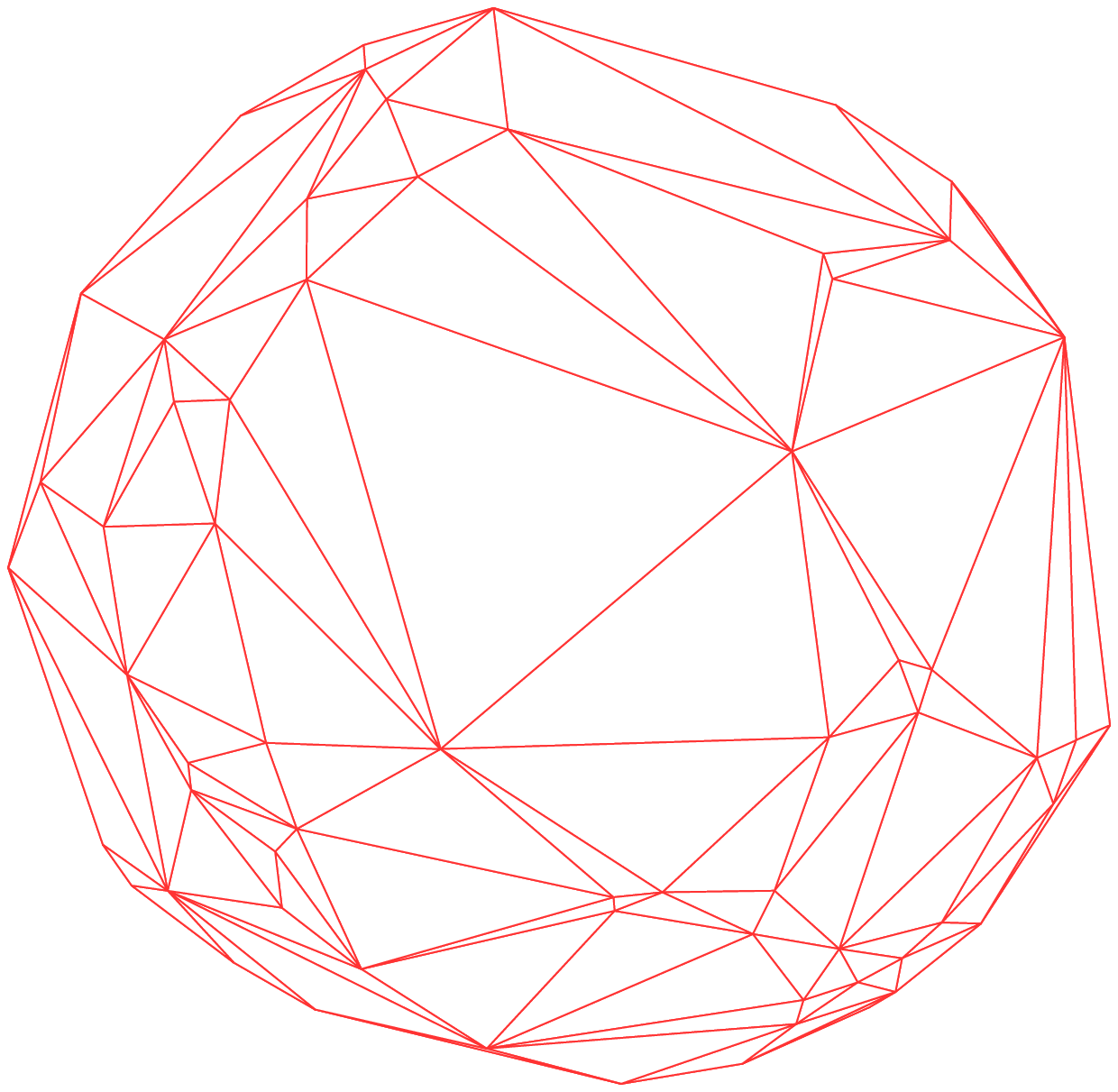}
\epsfxsize=2.3cm\epsfbox{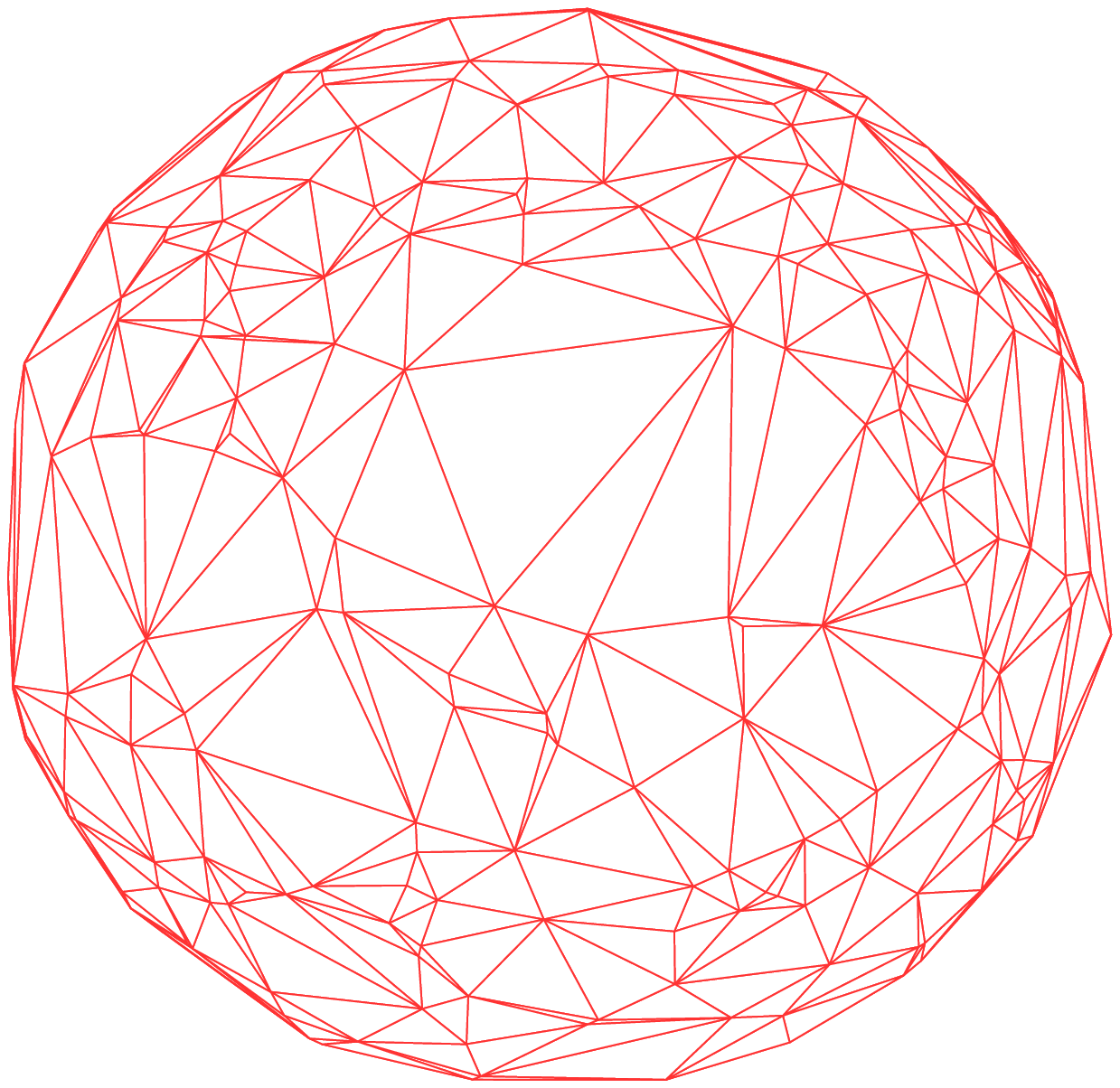}
\epsfxsize=2.3cm\epsfbox{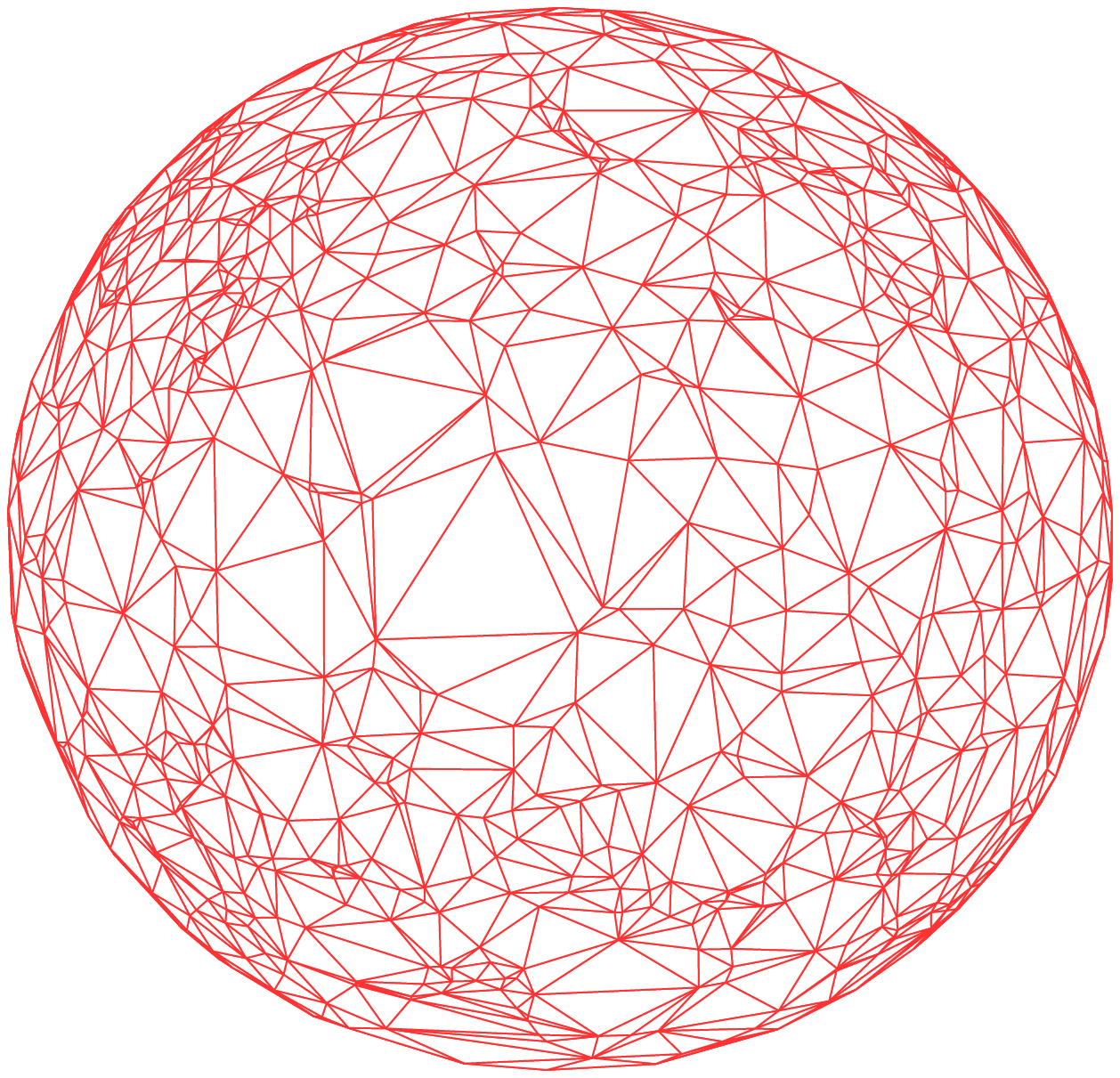}
\epsfxsize=2.3cm\epsfbox{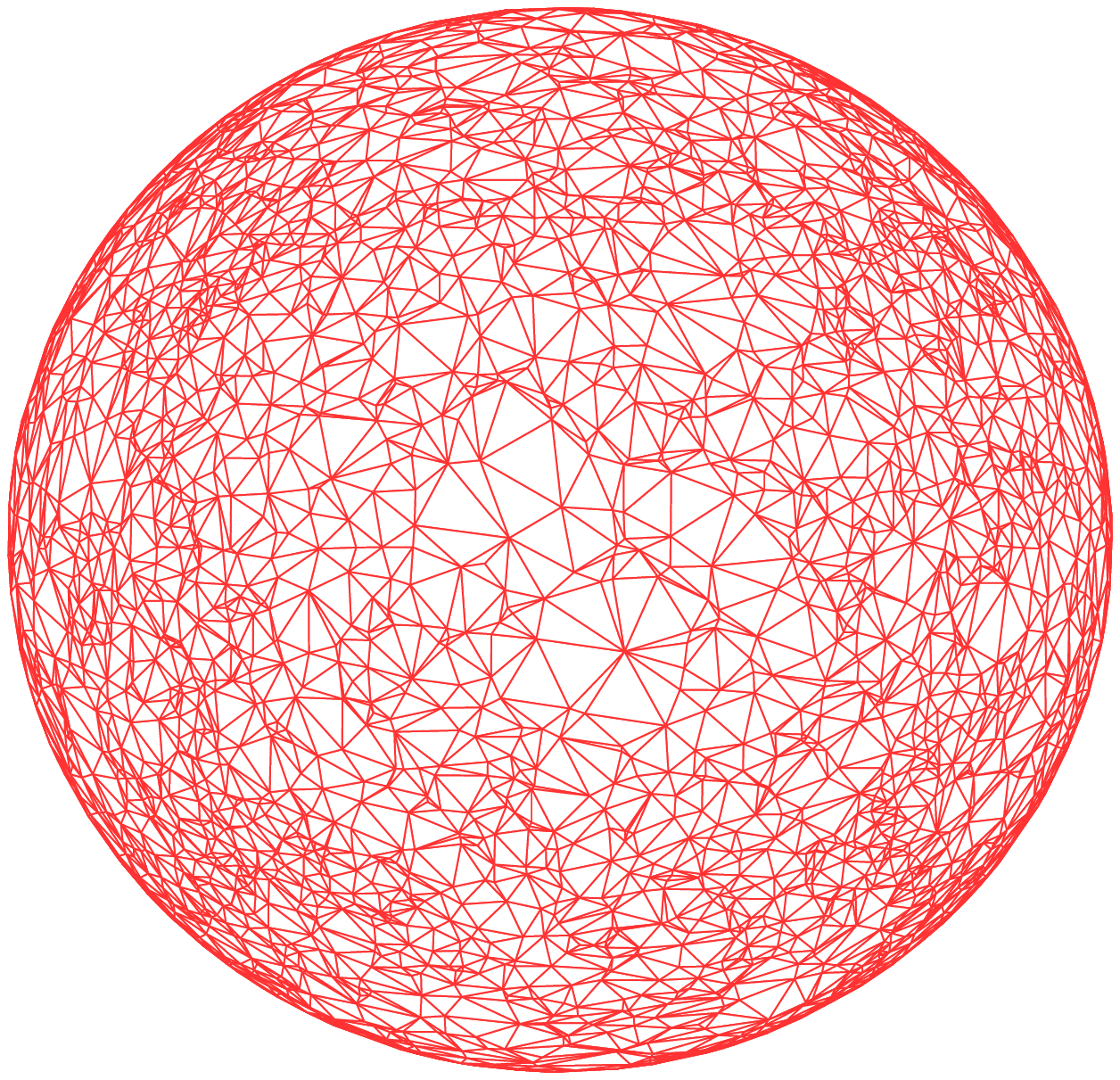}
\end{center}
{\normalsize {\bf Fig. 3.} Our test random triangulations. From
left to right, the number of vertices is 30, 100, 400, 1300, 5000
respectively. }

From these numerical results, we can draw the following
conclusions:

1. For the regular vertices with the valence greater than 4, or the
umbilical  points, the discrete scheme $G^{(5)}$ converges to the
real Gaussian curvature. This agrees with the theoretical result.

2. At the regular vertices and the umbilical points, the difference
between $G^{(2)}$ and $G^{(4)}$ is very small.

\bigskip

\noindent {\bf Acknowledgments.} Part of work is finished when the
first author visits Technical University of Berlin in 2007-08.
Zhiqiang Xu is Supported by the NSFC grant 10401021 and a Sofia
Kovalevskaya prize awarded to Olga Holtz. Guoliang Xu is supported
by NSFC grant 60773165 and National Key Basic Research Project of
China (2004CB318000).

\end{document}